\documentclass{article}
\usepackage{spconf,amsmath,graphicx}

\usepackage{url,amsfonts,enumitem,booktabs,subfigure,multirow}
\usepackage{algorithm}
\usepackage{algorithmic}

\title{Learning the Dynamic Correlations and Mitigating Noise by Hierarchical Convolution for Long-term Sequence Forecasting}

\name{Zhihao Yu$^{1,2}$, Liantao Ma$^{2,3,*}$, Yasha Wang$^{2,3,*}$, Junfeng Zhao$^{2,3}$ \thanks{*Corresponding author.}}
\address{$^{1}$School of Computer Science, Peking University, Beijing, China\\
$^{2}$Key Laboratory of High Confidence Software Technologies, Ministry of Education, Beijing, China\\
$^{3}$National Engineering Research Center of Software Engineering, Peking University, Beijing, China\\
}
\begin{document}
\ninept
\maketitle
\begin{abstract}
Deep learning algorithms, especially Transformer-based models, have achieved significant performance by capturing long-range dependencies and historical information. However, the power of convolution has not been fully investigated. Moreover, most existing works ignore the dynamic interaction among variables and evolutionary noise in series. Addressing these issues, we propose a \textbf{H}ierarchical \textbf{M}emorizing \textbf{Net}work (HMNet).
In particular, a hierarchical convolution structure is introduced to extract the information from the series at various scales.
Besides, we propose a dynamic variable interaction module to learn the varying correlation and an adaptive denoising module to search and exploit similar patterns to alleviate noises.
These modules can cooperate with the hierarchical structure from the perspective of fine to coarse grain. Experiments on five benchmarks demonstrate that HMNet significantly outperforms the state-of-the-art models by 10.6\% on MSE and 5.7\% on MAE. Our code is released at \url{https://github.com/yzhHoward/HMNet}.
\end{abstract}
\section{Introduction}

Multivariate long-term sequence forecasting (LTSF) plays a critical role in predicting changes over a long period in the future based on the historical data of the series. 
Forecasting the changes has dramatically impact on our modern societies in various domains, such as energy \cite{somu2021deep}, transportation \cite{lan2022dstagnn}, and finance \cite{imajo2021deep}. 


Traditionally, statistical methods such as ARIMA \cite{box1968some}, VAR \cite{litterman1986forecasting}, and Gaussian process models \cite{frigola2013bayesian} are used in time series forecasting. However, these methods fall short in making accurate predictions for LTSF, which is more challenging.
Recently, numerous deep-learning-based models try to solve these challenges, such as DeepAR \cite{salinas2020deepar} and TCN \cite{bai2018empirical}. Following the success of Transformer \cite{vaswani2017attention} in natural language processing tasks, various works \cite{li2019enhancing,tang2021probabilistic,zhou2021informer,liu2022pyraformer} use attention to capture long-range dependencies between time steps.


Despite the success, the effectiveness of convolution structure has not been fully exploited. As \cite{zeng2023transformers} showing that Transformers are not such effective, \cite{liu2022time} extracts at multiple resolutions with dilated CNNs and performs better than Transformers. However, as depicted in \cite{zeng2023transformers}, complicated models not always bring better performance.

Moreover, the previous works overlook the fact that correlations across variables tend to change over time. Variables may exhibit relevance at certain periods but not at others. 
Furthermore, correlations among variables can also change at different scales. Previous works encode variables via time-invariant approaches, which prevent capturing these dynamic correlations, let alone across different scales.



Furthermore, most methods ignore the negative impact of noise on time series forecasting. 
Noise can deceive the model and introduce spurious correlations \cite{calude2017deluge}, which could become particularly pronounced when learning dynamic correlations. In general, Fourier transform and wavelet transform have been employed to remove specific spectra \cite{zhou2022fedformer,zhou2022film} or spectras below a threshold value \cite{boto2010wavelet,xu2018wavelet} to reduce noise. However, these approaches are not effective in handling all cases. Additionally, the noise intensity can be evolutionary in the series. Therefore, a flexible denoising mechanism is in need.


In this paper, we propose a framework named \textbf{H}ierarchical \textbf{M}emorizing \textbf{Net}work (HMNet) to learn correlations and denoise dynamically with simple but effective convolution structure for LTSF. HMNet first embeds the variables separately, then adopts multi-level Memorizing Convolution Blocks (MC-Blocks) to explore the dynamic temporal correlations among variables and denoise by similar patterns in the dataset. Cooperating with a hierarchical architecture which extract the multi-resolution representations, HMNet captures correlations and performs denoising from multiple perspectives.
The key contributions of this work are as follows:

\begin{itemize}[leftmargin=*]
\item We propose a novel convolution based hierarchical architecture to extract the multi-resolution representations from series.
\item We design a dynamic variable interaction to explore the correlations across variables in different scales.
\item We identify and exploit similar patterns in the sequence hierarchically to flexibly alleviate the evolutionary noise.
\item We conduct extensive experiments on five benchmark datasets, demonstrating that the proposed HMNet outperforms state-of-the-art methods by 10.6\% in MSE and 5.7\% in MAE. Besides, HMNet exhibits robust performance in the presence of noise.
\end{itemize}

\begin{figure*}[h]
  \centering
  \subfigure[MC-Block] {
  \label{framework:mcblock}
  \includegraphics[width=0.29\linewidth]{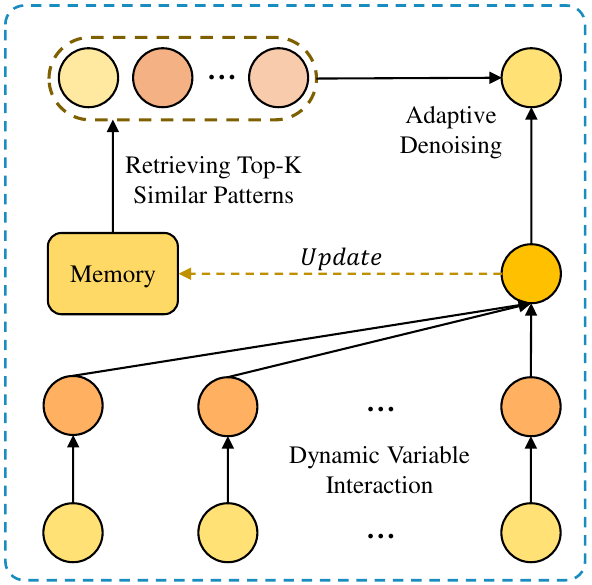}  
  }
  \subfigure[HMNet] {
  \label{framework:full}
  \includegraphics[width=0.60\linewidth]{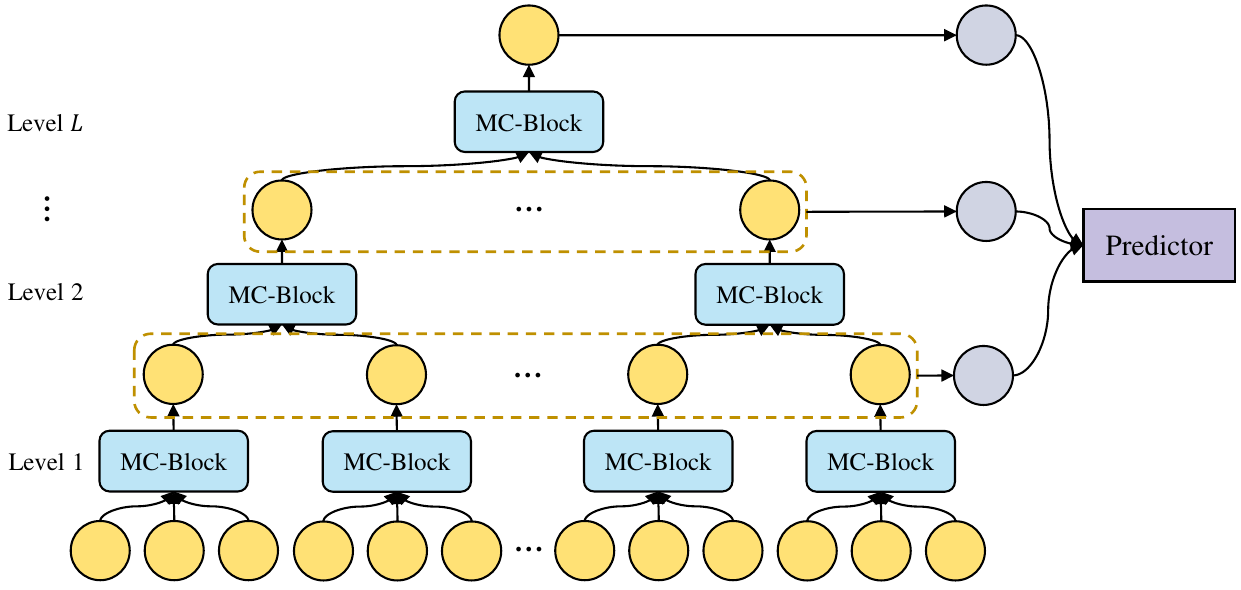}  
  }
  \vspace{-6pt}
  \caption{The framework of the Hierarchical Memorizing Network (HMNet). The HMNet consists of hierarchical Memorizing Convolution Blocks (MC-Blocks). The parameters and memory in the MC-Blocks at the same level are shared}
  \label{framework}
  \vspace{-6pt}
\end{figure*}

\section{Preliminary}
Formally, the multivariate time series data $x \in \mathbb{X}$ is defined as: given a series of $N$ variables with $T$ observations, then the record of $n$-th variable observed at $t$ can be represented by $x_{t, n}$, where $1 \le t \le T$ and $1 \le n \le N$. The time series forecasting problem aims to predict the future $H$ steps $\hat{x} = x_{T+1:T+H, n}$ according to $x_{1:T, n}$ and its timestamp, where $1 \le n \le N$. Long-term sequence forecasting encourages a longer horizon of the output sequence.

\section{Hierarchical Memorizing Network}
In this section, we present the proposed HMNet. As depicted in Figure \ref{framework:full}, HMNet consists of hierarchical memorizing convolution blocks (MC-Blocks) and a predictor. Each MC-Block comprises dynamic variable interaction, convolution unit, adaptive denoising, and memory, which is illustrated in Figure \ref{framework:mcblock}. HMNet is capable of extracting representations from different scales, capturing the evolving correlations among variables, and performing denoising based on similar patterns.
The predictor is responsible for integrating representations from every scale and forecasting with the given horizon.


\subsection{Variable-Specific Embedding}
%
We first discuss the variable-specific embedding in HMNet.
Most deep learning methods project all the variables into a shared vector, then learn temporal correlations and forecast these variables. However, this paradigm forces the learned embedding parameters to capture an "average" representation that may not fully capture the specific characteristics of individual variables \cite{cirstea2022triformer}. 
To address this limitation, we encode each variable separately, allowing us to capture dynamic correlations more effectively in subsequent steps.
Formally, given a series $x \in \mathbb{R}^{T \times N}$ and its timestamp, we extract the latent representation $h \in \mathbb{R}^{T \times N \times d}$ by linear projections, where $d$ denotes the dimension of hidden size. 

\subsection{Memorizing Convolution Block}

We introduce Memorizing Convolution Blocks (MC-Blocks) as a key component for learning dynamic correlations and denoising the representations. To extract information and denoise from multiple scales, we employ the MC-Blocks hierarchically. The representation extracted from a lower-level block is passed to the subsequent higher-level block, allowing us to extract knowledge in a bottom-up manner. Each level represents the overall information of the sequence from a specific perspective. Additionally, the parameters and memory in the MC-Blocks at the same level are shared, enhancing the efficiency of information processing and memory utilization.


\subsubsection{Dynamic Variable Interaction}
The dynamic variable interaction module is responsible for extracting the dynamic correlations among variables. 
It operates at a minimum granularity of one time step. We leverage the hierarchical structure to learn correlations at steply, short-term, and long-term scales.
The correlation $v$ is calculated by $v = W_v h$ in the variable dimension and $W_v \in \mathbb{R}^{N \times N}$ is a parameter matrix. We restrict the diagonal of $W_v$ to be all zeros to suppress the information from the given variable and enhance the information from the others.

The correlations are mutable in the multivariate series, indicating a demand to flexibly fuse the feature correlations $v$ with the representation $h$ instead of simply superimposing them.
Thus, we integrate them by an adaptive parameter $\alpha \in \mathbb{R}^N$ to determine the weight of information between them as described in Equation \ref{eq:alpha}.
\begin{equation}
\label{eq:alpha}
\begin{split}
\alpha &= \sigma(W_{\alpha}(h | v)),\\
h_v &= \alpha * h + (1 - \alpha) * v.
\end{split}
\end{equation}
$(\cdot | \cdot)$ denotes the concatenation of two matrices, $W_\alpha \in \mathbb{R}^{2N \times N}$ is a learnable parameter, and $\sigma$ represents the sigmoid function.

\subsubsection{Convolution Unit}
Different from traditional works that rely on one-dimensional convolution \cite{bai2018empirical,sen2019think} scanning the same time step multiple times, we introduce a convolution unit that divides the sequence into blocks and aggregates them for computational efficiency. This approach allows us to accelerate the processing while merging information from multiple steps. Specifically, we adopt a one-dimensional convolution unit with the same receptive field and stride. Each variable is embed separately. 
For an embeded multivariate sequence $h_v \in \mathbb{R}^{T \times N \times d}$, the convolutional unit divides it into $T/S$ blocks and generates latent feature $h_c \in \mathbb{R}^{T/S \times N \times d}$. 

\subsubsection{Adaptive Denoising}
To mitigate the noise, we employ an adaptive denoising approach that leverages similar patterns in the representations.
For retrieving similar patterns, one straightforward way is to search for patterns in the sequence or the mini-batch \cite{zhang2021grasp}. However, finding appropriate patterns in a small field can be challenging. Therefore, we propose a memory that stores representations of patterns. Nevertheless, memorizing the entire dataset brings excessive memory and query time overhead for large datasets.
To contend with this problem, we implement a memory with a fixed size and incorporate a first-in-first-out mechanism to balance performance and efficiency.


We employ memory to store the aggregated features obtained from the convolution unit. Each level of the hierarchy has its own independent memory. To simplify the pattern extraction process, we do not back-propagate gradients into the memory, thereby reducing computational effort. As model updates, there exists a distributional shift for patterns originate from different epochs. Especially for large memory, older patterns may become stale. We normalize the patterns to ensure that they are of the same magnitude.


For retrieving and exploiting representations in the memory, we search top-$K$ patterns which are most similar to the normalized aggregated feature and obtain a set of similar patterns $s$. 
Next, we calculate the similarity coefficient $\kappa$ between the normalized feature $h_c$ and similar patterns. By applying $\kappa$, we effectively prioritize the most relevant and informative patterns in the denoising process.
\begin{equation}
\kappa(h_c, s) = \frac{\mathrm{exp}(V_{\kappa} h_c (W_{\kappa} s)^\top / \sqrt{d})}{\sum_j^K \mathrm{exp}(V_{\kappa} h_c (W_{\kappa} s_j)^\top)}.
\end{equation}
The fused representation $h_s$ of similar patterns is obtained by:
\begin{equation}
h_s = \kappa(h_c, s) \cdot U_{\kappa} s,
\end{equation}
where $U_{\kappa}, V_{\kappa}, W_{\kappa} \in \mathbb{R}^{d\times d}$ are projection matrices.

The noise is averaged and smoothed since $h_s$ contains information from similar patterns. 
Owing to the fact that the amount of noise in sequences are different, we apply adaptive integration. The calculation of denoised representation $h_d$ can be formalized as:

\begin{equation}
\begin{split}
\beta &= \sigma(W_{\beta}(h_c | h_s)), \\
h_d &= \beta * h_c + (1 - \beta) * h_s. \\
\end{split}
\end{equation}



\subsection{Predictor}
With the denoised representation $h_d$ from MC-Blocks, we embed them separately by level. The overall representation $f_l$ of level $l$ with $P$ MC-Blocks is defined as 

\begin{equation}
f_l = W_l (h_{d, 1} | h_{d, 2} | ... | h_{d, P}).
\end{equation}

Then we employ representations from all levels to forecast since it provides more information and makes optimizing the lower levels easier. The forecasted series $\hat{x}$ can be computed by

\begin{equation}
\hat{x} = \mathrm{MLP} (f_1 + f_2 + ..., f_L),
\end{equation}
where $L$ denotes the number of levels in HMNet.





\section{Experiments}
In this section, we perform forecasting experiments on real-world datasets and compare the latest state-of-the-art methods.

\subsection{Evaluation Settings}
We evaluated our model on electricity load (ETTm2), electricity consumption, exchange, traffic, and weather dataset following the experiment protocols of \cite{zhou2022fedformer,wu2021autoformer}.
The input length is fixed to 96, and the prediction lengths are fixed to be 96, 192, 336, and 720, respectively. 
We assess the MSE and MAE on these datasets.
HMNet is compared with FiLM \cite{zhou2022film}, SCINet \cite{liu2022time}, FEDformer \cite{zhou2022fedformer}, DLinear \cite{zeng2023transformers},  Autoformer \cite{wu2021autoformer}, and Triformer \cite{cirstea2022triformer}.

We utilize three levels of MC-Blocks with block sizes of 6, 4, and 4 from lower level to higher level. The number of the retrieved similar patterns $K$ is 16 and the memory size is 4096. The data normalization \cite{kim2022reversible,liu2022non} is adopted for stabilizing. We employ Faiss \cite{johnson2019billion} to implement fast retrieving of similar patterns. Hyper-parameters are used to determine whether to enable dynamic variable interaction and adaptive denoising for each level to improve flexibility.

\subsection{Result and Analysis}
The results of baselines and HMNet are reported in Table \ref{result}.
HMNet consistently achieves the best performance across most settings and datasets. Compared to the best baseline FiLM, HMNet brings 10.6\% on MSE and 5.7\% on MAE improvement on average. Notably, on the Traffic dataset, which consists of 862 variables, HMNet achieves a remarkable improvement of more than 16\% on MSE. Both SCINet and Triformer are hierarchical structures, neither of them performs as well as HMNet. FiLM and SCINet also adopt normalization and denormalization methods. However, they cannot capture the multi-scale dependencies in time series. 
Though DLinear outperforms FEDformer when its input length is 336 \cite{zeng2023transformers}, it does not perform well in the fair setting where the input length is 96.


\subsubsection{Ablation Study}
We perform ablation studies on the ECL and the Weather dataset to measure the impact of the proposed modules. 
We first remove both the dynamic variable interaction and adaptive denoising to evaluate the performance of the proposed hierarchical convolutional structure, namely as \textit{w/o I\&D}. The results in Table \ref{ablation} reveal that this architecture significantly surpasses baselines, justifying the effectiveness.
Besides, we remove the dynamic variable interaction and adaptive denoising modules individually, referred as \textit{w/o Interact} and \textit{w/o Denoise}. 
In conjunction with the hierarchical architecture, these two modules efficaciously discover dynamic correlations and alleviate evolutionary noises respectively from the time step to the scale of the entire sequence. 
We observe that both modules are beneficial in forecasting. The dynamic variable interaction module leads to an average improvement of 2.3\% on MSE, while the adaptive denoising module yields an average improvement of 2.2\%.

\begin{table}[h]
  \centering
  \small
  \begin{tabular}{c|c|cccccc}
    \toprule
    \multicolumn{2}{c|}{Methods} & \multicolumn{2}{c|}{w/o I\&D} & \multicolumn{2}{c|}{w/o Interact} & \multicolumn{2}{c}{w/o Denoise}  \\
    \midrule
    \multicolumn{2}{c|}{Metric} & MSE & MAE & MSE & MAE & MSE & MAE\\
    \midrule
    \multirow{4}{*}{\rotatebox[origin=c]{90}{ECL}} 
    & 96 & 0.167 & 0.265 & 0.166 & 0.263 & 0.164 & 0.263\\
    & 192 & 0.182 & 0.275 & 0.180 & 0.273 & 0.175 & 0.271\\
    & 336 & 0.202 & 0.294 & 0.200 & 0.293 & 0.197 & 0.292\\
    & 720 & 0.246 & 0.335 & 0.243 & 0.329 & 0.245 & 0.334\\
    \midrule
    \multirow{4}{*}{\rotatebox[origin=c]{90}{Weather}} 
    & 96 & 0.170 & 0.215 & 0.164 & 0.208 & 0.168 & 0.213\\
    & 192 & 0.218 & 0.256 & 0.212 & 0.254 & 0.216 & 0.254\\
    & 336 & 0.275 & 0.299 & 0.273 & 0.296 & 0.271 & 0.297\\
    & 720 & 0.357 & 0.349 & 0.352 & 0.347 & 0.355 & 0.348\\
    \bottomrule
  \end{tabular}
  \vspace{-6pt}
  \caption{Ablation study of HMNet on ECL and Weather dataset}
  \label{ablation}
  \vspace{-12pt}
\end{table}

\begin{table*}[h]
  \centering
  \small
  \begin{tabular}{c|c|cccccccccccccc}
    \toprule
    \multicolumn{2}{c|}{Methods} & \multicolumn{2}{c|}{HMNet} & \multicolumn{2}{c|}{FiLM} & \multicolumn{2}{c|}{SCINet} & \multicolumn{2}{c|}{FEDformer} & \multicolumn{2}{c|}{DLinear} & \multicolumn{2}{c|}{Autoformer} & \multicolumn{2}{c}{Triformer}\\
    \midrule
    \multicolumn{2}{c|}{Metric} & MSE & MAE & MSE & MAE & MSE & MAE & MSE & MAE & MSE & MAE & MSE & MAE & MSE & MAE\\
    \midrule
    \multirow{4}{*}{\rotatebox[origin=c]{90}{ETTm2}}
    ~ & 96 & \textbf{0.170}  & \textbf{0.253}  & 0.184  & 0.267  & 0.185  & 0.269  & 0.190  & 0.283  & 0.199  & 0.297  & 0.259  & 0.333  & 0.280  & 0.350  \\
    ~ & 192 & \textbf{0.235}  & \textbf{0.295}  & 0.249  & 0.307  & 0.252  & 0.311  & 0.257  & 0.324  & 0.285  & 0.361  & 0.281  & 0.337  & 0.486  & 0.462  \\
    ~ & 336 & \textbf{0.295}  & \textbf{0.335}  & 0.310  & 0.344  & 0.314  & 0.350  & 0.324  & 0.362  & 0.388  & 0.430  & 0.353  & 0.382  & 0.722  & 0.557  \\
    ~ & 720 & \textbf{0.393}  & \textbf{0.393}  & 0.410  & 0.403  & 0.413  & 0.405  & 0.425  & 0.421  & 0.543  & 0.516  & 0.433  & 0.427  & 1.831  & 0.839  \\
    \midrule
    \multirow{4}{*}{\rotatebox[origin=c]{90}{ECL}}
    ~ & 96 & \textbf{0.162}  & \textbf{0.261}  & 0.199  & 0.276  & 0.192  & 0.298  & 0.183  & 0.302  & 0.224  & 0.316  & 0.201  & 0.318  & 0.245  & 0.339  \\
    ~ & 192 & \textbf{0.174}  & \textbf{0.270}  & 0.205  & 0.289  & 0.212  & 0.313  & 0.207  & 0.322  & 0.224  & 0.320  & 0.217  & 0.326  & 0.250  & 0.344  \\
    ~ & 336 & \textbf{0.195}  & \textbf{0.292}  & 0.218  & 0.312  & 0.232  & 0.331  & 0.212  & 0.327  & 0.237  & 0.334  & 0.240  & 0.344  & 0.266  & 0.359  \\
    ~ & 720 & \textbf{0.241}  & \textbf{0.330}  & 0.279  & 0.357  & 0.263  & 0.355  & 0.245  & 0.352  & 0.272  & 0.363  & 0.267  & 0.370  & 0.299  & 0.385  \\
    \midrule
    \multirow{4}{*}{\rotatebox[origin=c]{90}{Exchange}}
    ~ & 96 & \textbf{0.083}  & \textbf{0.200}  & 0.089  & 0.213  & 0.096  & 0.220  & 0.139  & 0.267  & 0.117  & 0.258  & 0.152  & 0.283  & 0.195  & 0.323  \\
    ~ & 192 & \textbf{0.173}  & \textbf{0.296}  & 0.182  & 0.305  & 0.185  & 0.309  & 0.276  & 0.382  & 0.206  & 0.347  & 0.319  & 0.410  & 0.406  & 0.476  \\
    ~ & 336 & \textbf{0.325}  & \textbf{0.412}  & 0.343  & 0.423  & 0.332  & 0.419  & 0.445  & 0.493  & 0.341  & 0.449  & 0.753  & 0.651  & 0.695  & 0.624  \\
    ~ & 720 & 0.869  & 0.701  & 0.918  & 0.720  & 0.875  & 0.706  & 1.152  & 0.829  & \textbf{0.817}  & \textbf{0.692}  & 1.215  & 0.847  & 1.229  & 0.874  \\
    \midrule
    \multirow{4}{*}{\rotatebox[origin=c]{90}{Traffic}}
    ~ & 96 & \textbf{0.486}  & \textbf{0.333}  & 0.582  & 0.356  & 0.630  & 0.419  & 0.576  & 0.359  & 0.754  & 0.471  & 0.627  & 0.375  & 0.758  & 0.426  \\
    ~ & 192 & \textbf{0.500}  & \textbf{0.341}  & 0.604  & 0.367  & 0.687  & 0.450  & 0.612  & 0.381  & 0.706  & 0.452  & 0.625  & 0.393  & 0.754  & 0.421  \\
    ~ & 336 & \textbf{0.524}  & \textbf{0.353}  & 0.614  & 0.372  & 0.730  & 0.470  & 0.624  & 0.387  & 0.713  & 0.455  & 0.633  & 0.375  & 0.768  & 0.426  \\
    ~ & 720 & \textbf{0.562}  & \textbf{0.375}  & 0.692  & 0.427  & 0.789  & 0.493  & 0.630  & 0.385  & 0.755  & 0.472  & 0.668  & 0.408  & 0.806  & 0.443  \\
    \midrule
    \multirow{4}{*}{\rotatebox[origin=c]{90}{Weather}}
    ~ & 96 & \textbf{0.159}  & \textbf{0.206}  & 0.195  & 0.238  & 0.174  & 0.220  & 0.245  & 0.330  & 0.199  & 0.258  & 0.259  & 0.333  & 0.170  & 0.235  \\
    ~ & 192 & \textbf{0.207}  & \textbf{0.249}  & 0.238  & 0.272  & 0.224  & 0.262  & 0.274  & 0.334  & 0.240  & 0.297  & 0.308  & 0.366  & 0.215  & 0.279  \\
    ~ & 336 & \textbf{0.267}  & \textbf{0.293}  & 0.288  & 0.305  & 0.281  & 0.303  & 0.333  & 0.372  & 0.286  & 0.335  & 0.374  & 0.4055  & 0.271  & 0.321  \\
    ~ & 720 & \textbf{0.346}  & \textbf{0.344}  & 0.359  & 0.351  & 0.357  & 0.352  & 0.409  & 0.419  & 0.348  & 0.384  & 0.422  & 0.428  & 0.349  & 0.380 \\
    \bottomrule
  \end{tabular}
  \caption{Performance comparison on datasets. The best results are highlighted in bold.}
  \label{result}
  \vspace{-6pt}
\end{table*}

\subsubsection{Effect of Denoising}
We add random Gaussian noise to the dataset to further explore the denoising performance of HMNet and compare with previous methods. We investigate two settings by applying random Gaussian noise to all variables in the input series with a given probability: (1) $\mathcal{N}(0, 1)$ to residual components, and (2) $\mathcal{N}(1, 1)$ to both the trend and residual. These settings represent scenarios where noise affects different parts. Note that this experiment aims to verify the model's ability to resist noise rather than to forecast the noise, so the noise is not added to the target sequence. We evaluate the performance on the ECL dataset with noise probabilities ranging from 0 to 0.4.

\begin{figure}[h]
  \centering
  \subfigure[$\mathcal{N}(0, 1)$] {
    \label{noise:ecl}
    \includegraphics[width=0.47\columnwidth]{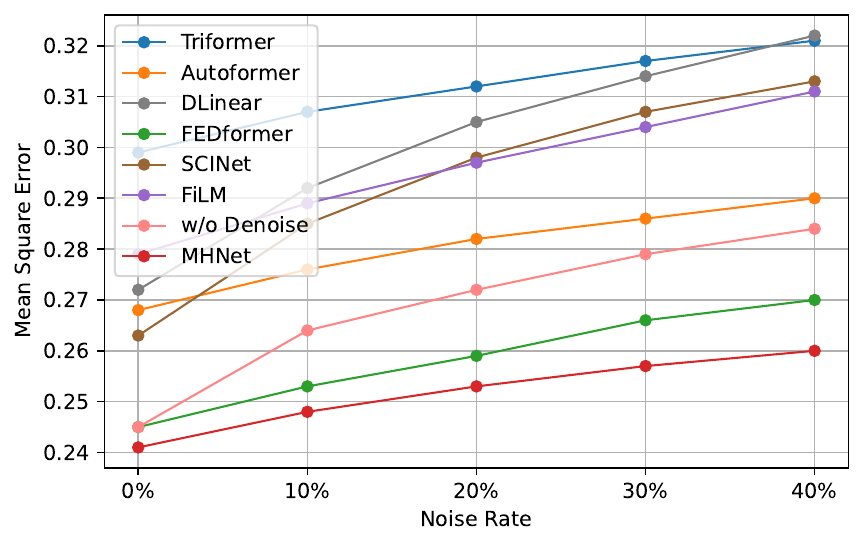}
  }
  \subfigure[$\mathcal{N}(1, 1)$] {
    \label{noise:weather}
    \includegraphics[width=0.47\linewidth]{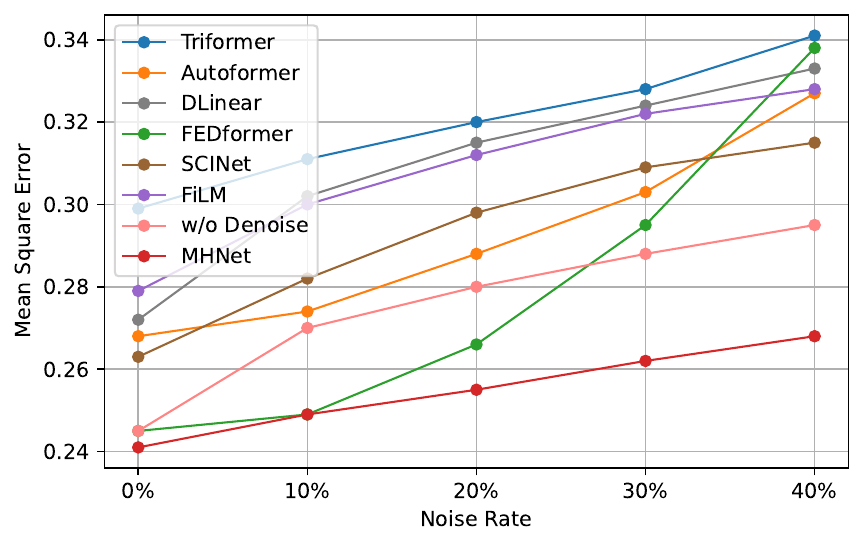}
  }
  \vspace{-6pt}
  \caption{The performance of models on the ECL dataset with Gaussian noise. The forecasted length is 720}
  \vspace{-12pt}
  \label{noise}
\end{figure}

By observing the results in Figure \ref{noise:ecl} and Figure \ref{noise:weather}, we notice that the performance decay of HMNet is the slightest as the increment of noise rate on both settings. DLinear and SCINet are the two most sensitive models to noise from residual disturbances, while FEDformer and Autoformer perform poorly when noise affects both the trend and residual components. The encoder-decoder structure in FEDformer and Autoformer provides accurate predictions, whereas it is more sensitive to noise. Although FiLM claims to reduce noise for robust forecasting, it does not perform well in our experimental settings. Notably, there is a significant performance gap between HMNet and the \textit{w/o Denoise} variant, which widens as the noise ratio increases. This indicates that the adaptive denoising module in HMNet is effective, especially when dealing with more noises.

\subsubsection{Study on Memory Size and Amount of Similar Patterns}
We investigate the impact of varying the memory size $M$ and the number $K$ of retrieved similar patterns in HMNet. We try different combinations of memory size and similar pattern to evaluate their importance. From the results presented in Table \ref{memsize}, HMNet is not sensitive to changes in memory size and the number of similar patterns. Interestingly, even though we merely adopt small memory and select one similar pattern for the given sequence, there is still a considerable gain in performance compared to \textit{w/o Denoise}. Despite increasing the memory size and searching more patterns benefit the performance, the cost-effectiveness is relatively low. 
We conclude that the information in similar patterns is some of homogenous. Thus, the benefits of retrieving more patterns become less significant.
However, in situations where higher accuracy is required, using a larger memory size and retrieving more patterns is helpful.

\begin{table}[h]
  \centering
  \small
  \begin{tabular}{c|cccccc}
    \toprule
    Methods & \multicolumn{2}{c|}{M=256 K=1} & \multicolumn{2}{c|}{M=4096 K=16} & \multicolumn{2}{c}{M=16384 K=64}\\
    \midrule
    Metric & MSE & MAE & MSE & MAE & MSE & MAE\\
    \midrule
    96 & 0.162 & 0.208 & 0.159 & 0.206 & 0.158 & 0.206\\
    192 & 0.209 & 0.252 & 0.207 & 0.249 & 0.206 & 0.248\\
    336 & 0.269 & 0.293 & 0.267 & 0.293 & 0.267 & 0.292\\
    720 & 0.348 & 0.346 & 0.346 & 0.344 & 0.344 & 0.343\\
    \bottomrule
  \end{tabular}
  \vspace{-6pt}
  \caption{Study on memory size and amount of retrieved similar patterns of HMNet on the Weather dataset. M denotes memory size}
  \label{memsize}
  \vspace{-12pt}
\end{table}

\section{Conclusion and Future Work}

In this paper, we propose a novel hierarchical memorizing network (HMNet) for long-term sequence forecasting. HMNet effectively extracts multi-level representation by hierarchical convolutional architecture.
The proposed dynamic variable interaction captures dynamic correlations among variable flexibly, while the adaptive denoising module leverages similar patterns to smooth and mitigate the evolutionary noise. 
The hierarchical architecture of HMNet allows for capturing correlations and denoising from multiple perspectives. 
We conduct extensive experiments to show that the proposed model achieves state-of-the-art performance with a significant improvement and exhibits robustness in the presence of noise.
In future work, we plan to explore ways to quantify the noise and leverage similar patterns with less noise to further enhance denoising.


\bibliographystyle{IEEEbib}
\bibliography{reference}

\end{document}